\documentclass[review]{cvpr}

\usepackage{times}
\usepackage{epsfig}
\usepackage{graphicx}
\usepackage{amsmath}
\usepackage{amssymb}
\usepackage[dvipsnames]{xcolor}

\usepackage[pagebackref=true,breaklinks=true,colorlinks,bookmarks=false]{hyperref}

\toggletrue{cvprfinal}
\def\confYear{CVPR 2021}

\begin{document}

\newcommand{\netex}{NeuTex}

\title{\netex: Neural Texture Mapping for Volumetric Neural Rendering}

\author{
  {Fanbo Xiang$^1$, Zexiang Xu$^2$, Milo\v{s} Ha\v{s}an$^2$, Yannick Hold-Geoffroy$^2$, Kalyan Sunkavalli$^2$, Hao Su$^1$}\\\\
  {$^1$ University of California, San Diego}\\
  {$^2$ Adobe Research}
}

\maketitle

\newcommand{\Comment}[1]{}

\newcommand{\boldstart}[1]{\noindent\textbf{#1}}
\newcommand{\boldstartspace}[1]{\vspace{0.1in}\noindent\textbf{#1}}

\newcommand{\PixelColor}{\mathbf{I}}
\newcommand{\PixelMask}{M}
\newcommand{\Color}{\mathbf{c}}
\newcommand{\Pos}{\mathbf{x}}
\newcommand{\Dir}{\mathbf{d}}
\newcommand{\UV}{\mathbf{u}}
\newcommand{\Normal}{\mathbf{n}}
\newcommand{\Refl}{\mathbf{r}}

\newcommand{\Trans}{T}
\newcommand{\Dens}{\sigma}
\newcommand{\Step}{\delta}
\newcommand{\Func}{F}
\newcommand{\FuncDens}{F_\Dens}
\newcommand{\FuncUV}{F_\text{uv}}
\newcommand{\FuncInv}{F^{-1}_\text{uv}}
\newcommand{\FuncTex}{F_\text{tex}}
\newcommand{\BlendW}{w}

\begin{abstract}
Recent work \cite{mildenhall2020nerf,bi2020deep} has demonstrated that volumetric scene representations combined with differentiable volume rendering can enable photo-realistic rendering for challenging scenes that mesh reconstruction fails on. 
However, these methods entangle geometry and appearance in a ``black-box'' volume that cannot be edited. 
Instead, we present an approach that explicitly disentangles geometry---represented as a continuous 3D volume---from appearance---represented as a continuous 2D texture map.
We achieve this by introducing a 3D-to-2D texture mapping (or surface parameterization) network into volumetric representations.
We constrain this texture mapping network using an additional 2D-to-3D inverse mapping network and a novel cycle consistency loss to make 3D surface points map to 2D texture points that map back to the original 3D points.  
We demonstrate that this representation can be reconstructed using only multi-view image supervision and generates high-quality rendering results. 
More importantly, by separating geometry and texture, we allow users to edit appearance by simply editing 2D texture maps.

 

\end{abstract}

\section{Introduction}
{\let\thefootnote\relax\footnote{{Research partially done When F. Xiang was an intern at Adobe Research.}}}

Capturing and modeling real scenes from image inputs is an extensively studied problem in vision and graphics. 
One crucial goal of this task is to avoid the tedious manual 3D modeling process and directly provide a renderable and editable 3D model that can be used for realistic rendering in applications, like e-commerce, VR and AR. 
Traditional 3D reconstruction methods \cite{schoenberger2016sfm,schoenberger2016mvs,kazhdan2006poisson} usually reconstruct objects as meshes. 
Meshes are widely used in rendering pipelines; they are typically combined with mapped textures for appearance editing in 3D modeling pipelines.

However, mesh-based reconstruction is particularly challenging and often cannot synthesize highly realistic images for complex objects. 
Recently, various neural scene representations have been presented to address this scene acquisition task. 
Arguably the best visual quality is obtained by approaches like NeRF \cite{mildenhall2020nerf} and Deep Reflectance Volumes \cite{bi2020deep} that leverage differentiable volume rendering (ray marching).
However, these volume-based methods do not (explicitly) reason about the object's surface and entangle both geometry and appearance in a volume-encoding neural network. 
This does not allow for easy editing---as is possible with a texture mapped mesh---and significantly limits the practicality of these neural rendering approaches.

\begin{figure}[t]
\includegraphics[width=\linewidth]{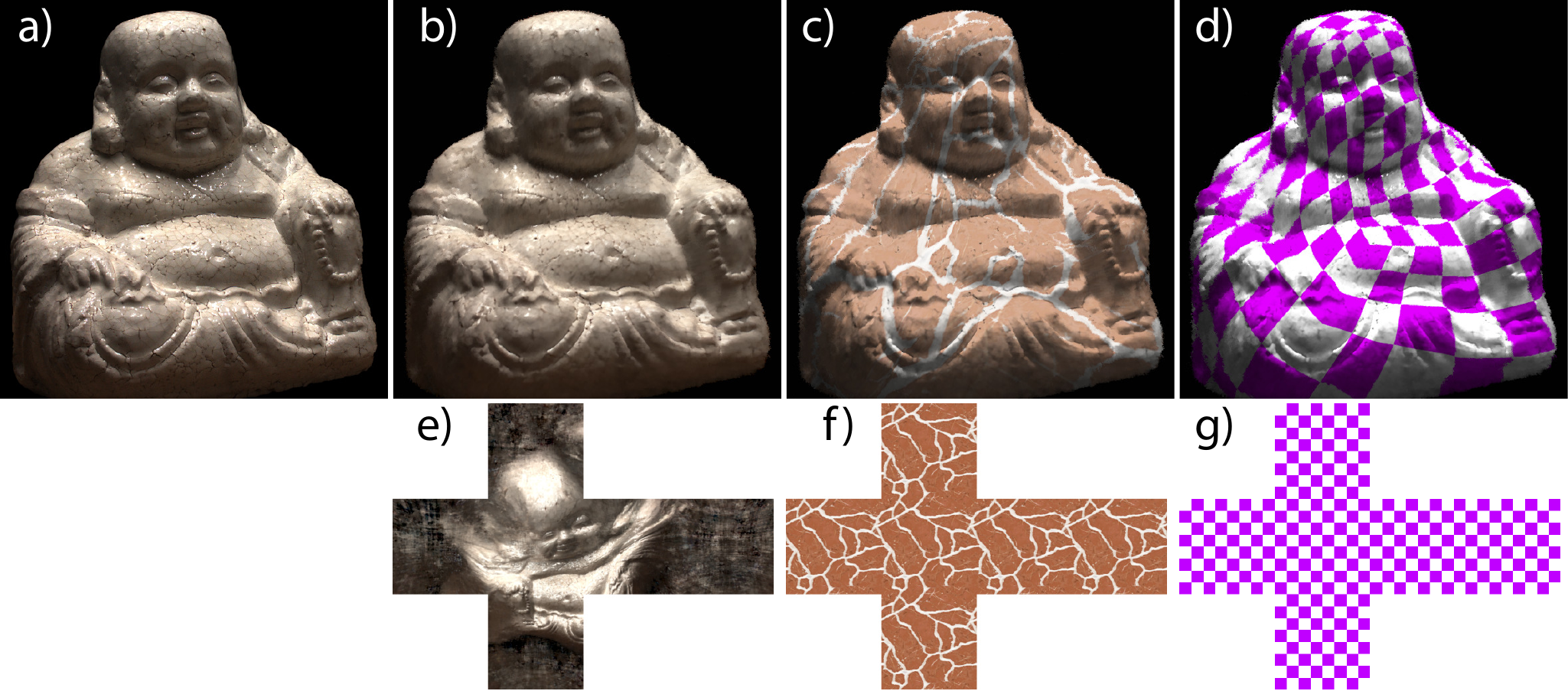}
    \caption{
    \netex{} is a neural scene representation that represents geometry as a 3D volume 
    but appearance as a 2D neural texture in an automatically discovered texture UV space, shown as a cubemap in (e).
    \netex{} can synthesize highly realistic images (b) that are very close to the ground-truth (a).
    Moreover, it enables intuitive surface appearance editing directly in the 2D texture space; 
    we show an example of this in (c), by using a new texture (f) to modulate the reconstructed texture.
    Our discovered texture mapping covers the object surface uniformly, as illustrated in (d), by rendering the object using a uniform checkerboard texture (g). 
    }
    \label{fig:teaser}
\end{figure}

Our goal is to make volumetric neural reconstruction more practical by enabling both realistic image synthesis and flexible surface appearance editing. 
To this end, we present \netex---an approach that explicitly disentangles scene geometry from appearance.
\netex{} represents geometry with a volumetric representation (similar to NeRF) but \emph{represents surface appearance using 2D texture maps}.
This allows us to leverage differentiable volume rendering to reconstruct the scene from multi-view images, while allowing for conventional texture-editing operations (see Fig.~\ref{fig:teaser}).

As in NeRF \cite{mildenhall2020nerf}, we march a ray through each pixel, regress volume density and radiance (using fully connected MLPs) at sampled 3D shading points on the ray, accumulate the per-point radiance values to compute the final pixel color.
NeRF uses a single MLP to regress both density and radiance in a 3D volume.
While we retain this volumetric density-based representation for geometry, \netex{} represents radiance in a 2D (UV) texture space.
In particular, we train a \emph{texture mapping} MLP to regress a 2D UV coordinate at every 3D point in the scene, and use another MLP to regress radiance in the 2D texture space for any UV location.
Thus, given any 3D shading point in ray marching, our network can obtain its radiance by sampling the reconstructed neural texture at its mapped UV location. 

Naively adding a texture mapping network to NeRF (and supervising only with a rendering loss) leads to a degenerate texture mapping that does not unwrap the surface and cannot support texture editing (see Fig.~\ref{fig:inverse}). 
To ensure that the estimated texture space reasonably represents the object's 2D surface, we introduce a novel cycle consistency loss. 
Specifically, we consider the shading points that contribute predominantly to the pixel color along a given ray, and correspond to the points either on or close to the surface. 
We train an additional \emph{inverse mapping} MLP to map the 2D UV coordinates of these high-contribution points \emph{back} to their 3D locations. 
Introducing this inverse-mapping network forces our model to learn a consistent mapping (similar to a one-to-one correspondence) between the 2D UV coordinates and the 3D points on the object surface. 
This additionally regularizes the surface reasoning and texture space discovery process.
As can be seen in Fig.~\ref{fig:teaser}, our full model recovers a reasonable texture space, that can support realistic rendering similar to previous work while also allowing for intuitive appearance editing. 

Our technique can be incorporated into different volume rendering frameworks.
In addition to NeRF, we show that it can be combined with Neural Reflectance Fields \cite{bi2020neural} to reconstruct BRDF parameters as 2D texture maps (see Fig.~\ref{fig:refresults}), enabling both view synthesis and relighting.

Naturally, \netex{} is more constrained than a fully-volumetric method; this leads to our final rendering quality to be on par or slightly worse than NeRF \cite{mildenhall2020nerf}. 
Nonetheless, we demonstrate that our approach can still synthesize photo-realistic images and significantly outperform both traditional mesh-based reconstruction methods \cite{schoenberger2016mvs} and previous neural rendering methods \cite{sitzmann2019scene,sitzmann2019deepvoxels}.
Most importantly, our work is the first to recover a meaningful surface-aware texture parameterization of a scene and enable surface appearance editing applications (as in Fig.~\ref{fig:teaser} and \ref{fig:results}).
This, we believe, is an important step towards making neural rendering methods useful in 3D design workflows.


\Comment{Modeling 3D scenes from captured images is a classical graphics and vision problem. Recently, various neural scene representations have been presented to address this problem. Differentiable ray-marching based approaches\cite{mildenhall2020nerf,lombardi2019neural,bi2020deep} can capture high-quality geometry and appearance, enabling applications such as view synthesis and relighting. These methods focus on training a per-scene network that encapsulates both the geometry and appearance in a point-wise volumetric representation, making it hard to do any appearance editing.

On the other hand, multi-view stereo methods, such as COLMAP\cite{schoenberger2016sfm,schoenberger2016mvs} uses matching correspondence to reconstruct surface point clouds. However, they often struggle with highly specular surfaces where correspondences are hard to find. Moreover, since these methods are not trained with rendering loss, it is harder for re-rendered images to match the original scenes.

A key motivation of this work is that we want to produce editable textures while preserving high-quality renderings.} 

\section{Related Work}
\boldstartspace{Scene representations.}
Deep learning based methods have explored various classical scene representations, including volumes \cite{ji2017surfacenet,wu20153d,qi2016volumetric,sitzmann2019deepvoxels}, point clouds \cite{qi2017pointnet,achlioptas2018learning,wang2018mvpnet}, meshes \cite{kanazawa2018learning,wang2018pixel2mesh}, depth maps \cite{liu2015learning,huang2018deepmvs} and implicit functions \cite{chen2018learning,mescheder2018occupancy,niemeyer2020differentiable,yariv2020multiview}.
However, most of them focus on geometry reconstruction and understanding and do not aim to perform realistic image synthesis.
We leverage volumetric neural rendering \cite{mildenhall2020nerf,bi2020neural} for realistic rendering; our method achieves higher rendering quality than other neural rendering methods \cite{sitzmann2019deepvoxels,sitzmann2019scene}. 

\boldstartspace{Mesh-based reconstruction and rendering.}
3D polygonal meshes are one of the most popular geometry representations, widely used in 3D modeling and rendering pipelines.
Numerous traditional 3D reconstruction techniques have been proposed to 
directly reconstruct a mesh from multiple captured images, including structure from motion \cite{schoenberger2016sfm}, multi-view stereo \cite{furukawa2009accurate,kutulakos2000theory,schoenberger2016mvs}, and surface extraction \cite{lorensen1987marching,kazhdan2006poisson}.
Recently, many deep learning based methods \cite{vijayanarasimhan2017sfm,yao2018mvsnet,tang2018ba,chen2019point,cheng2020deep} have also been proposed, improving the reconstruction quality in many of these techniques.
In spite of these advances, it is still challenging to reconstruct a mesh that can directly be used to synthesize photo-realistic images.
In fact, many image-based rendering techniques \cite{buehler2001unstructured,bi2017patch,hedman2018deep} have been presented to fix the rendering artifacts from mesh reconstruction;
however, they often leverage view-dependent texture maps \cite{debevec1998efficient}, which cannot be easily edited.
We instead leverage volumetric neural rendering to achieve realistic image synthesis; our approach explicitly extracts surface appearance as view-independent textures, just like standard textures used with meshes, allowing for broad texture editing applications in 3D modeling and content generation.

\begin{figure*}[t]
    \centering
    \includegraphics[width=0.95\linewidth]{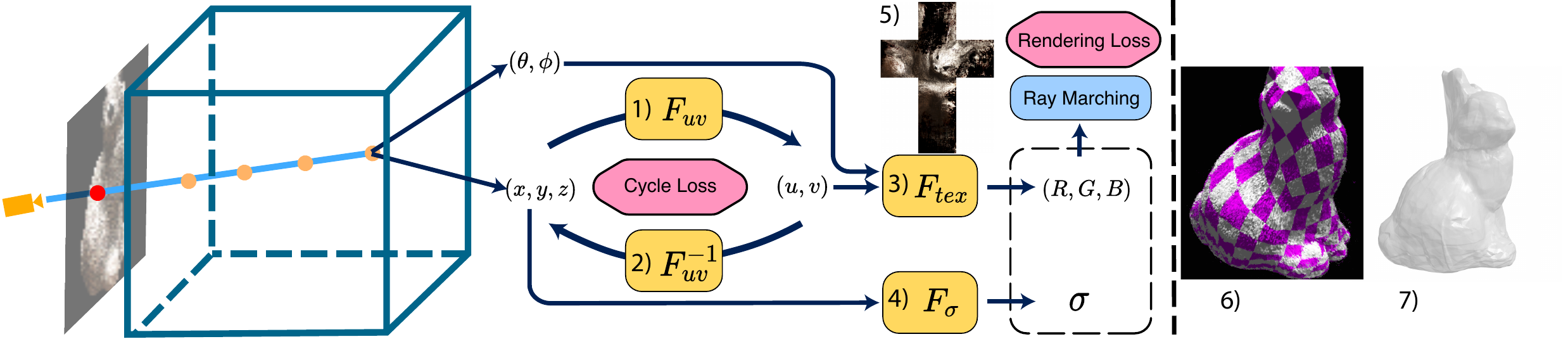}
    \caption{Overview.
    We present a disentangled neural representation consisting of multiple MLPs for neural volumetric rendering.
    As in NeRF \cite{mildenhall2020nerf}, for geometry we use an MLP (4) $\FuncDens$ to regress volume density $\Dens$ at any 3D point $\Pos=(x,y,z)$.
    In contrast, for appearance, we use a texture mapping MLP (1) $\FuncDens$  to map 3D points to 2D texture UVs, $\UV=(u,v)$, and a texture network (3) $\FuncTex$ to regress the 2D view-dependent radiance in the UV space given a UV $\UV$ and a viewing direction $\Dir=(\theta,\phi)$. 
    One regressed texture (for a fixed viewing direction) is shown in (5). 
    We also train an inverse mapping MLP (2) $\FuncInv$ that maps UVs back to 3D points.
    We leverage a cycle loss (Eqn.~\ref{eqn:cycleloss}) to ensure consistency between the 3D-to-2D mapping $\FuncUV$ and the 2D-to-3D $\FuncInv$ mapping at points on the object surface.
    This enables meaningful surface reasoning and texture space discovery, as illustrated by (6, 7).
    We demonstrate the meaningfulness of the UV space learned by $\FuncUV$ (6) by rendering the object with a uniform checkerboard texture.
    We also show the result of the inverse mapping network (7) by uniformly sampling UVs in the texture space and unprojecting them to 3D using $\FuncInv$, resulting in a reasonable mesh.
    }
    \label{fig:arch}
\end{figure*}

\boldstartspace{Neural rendering.}
Recently, deep learning-based methods have proposed to ameliorate or completely bypass mesh reconstruction to achieve realistic neural renderings of real scenes for view synthesis \cite{zhou2018stereo,xu2019deep,sitzmann2019scene,sitzmann2019deepvoxels}, relighting \cite{xu2018deep,philip2019multi,chen2020neural}, and many other image synthesis tasks \cite{lombardi2019neural}.
In particular, NeRF \cite{mildenhall2020nerf}, Deep Reflectance Volumes \cite{bi2020deep} and other relevant works \cite{bi2020neural,liu2020neural} model a scene using neural volumetric representations (that encode geometry and appearance) and leverage differentiable volume rendering \cite{max1995optical} to synthesize highly photo-realistic images.
However, these volume representations do not explicitly reason about the 2D surface of a scene and are essentially ``black-box'' functions that cannot be easily modified after reconstruction.
In contrast, we introduce a novel neural scene representation that offers direct access to the 2D surface appearance in volumetric neural rendering.
Our representation has disentangled geometry and appearance components, and models appearance as a 2D neural texture in a auto-discovered texture space.
Unlike previous volumetric neural rendering methods, this allows for easy texture/appearance editing.

\boldstartspace{Learning textures.}
Texture mapping is a standard technique widely used with meshes.
Here, surface appearance is represented by a 2D texture image and a 3D-to-2D mapping from every mesh vertex to the texture space. 
Textures can be easily controlled and edited by artists as needed to create diversities of scene appearance.
Recently, many deep learning based methods leverage texture-based techniques to model geometry or appearance in a scene \cite{henzler2020learning,groueix2018papier,thies2019deferred}.
Many works learn a 2D texture for a mesh by assuming a known mesh template \cite{kanazawa2018learning,saito2017photorealistic,goel2020shape}, focusing on reconstruction problems for specific object categories.
Our approach works for arbitrary shapes, and we instead learn a 2D texture in a volume rendering framework, discovering a 2D surface in the 3D volume space.  
Thies et al. \cite{thies2019deferred} optimize neural textures to do rendering for a known fixed mesh with given UV mapping.
In contrast, our approach simultaneously reconstructs the scene geometry as a volume, discovers a 2D texture UV space, and regresses a neural texture in the self-discovered texture space.
Other methods learn appearance by regressing colors directly from 3D points \cite{oechsle2019texture,oechsle2020learning}, which requires a known mesh and does not provide a 2D UV space necessary for texture editing.

AtlasNet \cite{groueix2018papier} and follow-up work \cite{poursaeed2020coupling} train neural networks to map 2D UV coordinates into 3D locations (like an inverse texture mapping), modeling an object 2D surface as an unwrapped atlas. 
These works focus on learning generalized geometry representations, and cannot be directly applied to arbitrary shapes or used for realistic rendering. 
Our network instead discovers a cycle mapping between a 2D texture space and a 3D volume, learning both a texture mapping and an inverse mapping.
We leverage differentiable volume rendering to model scene appearance from captured images for realistic rendering.
We also show that our neural texture mapping, supervised using a rendering loss and a cycle mapping loss, can discover a more uniform surface than a simple AtlasNet, supervised by a noisy point cloud from COLMAP \cite{schoenberger2016mvs} (see Fig.~\ref{fig:atlas}).




\section{Neural Texture Mapping}

\subsection{Overview}
We now present the \netex{} scene representation and demonstrate how to use it in the context of volumetric neural rendering.
While \netex{} can enable disentangled scene modeling and texture mapping for different acquisition and rendering tasks, in this section, we demonstrate its view synthesis capabilities with NeRF \cite{mildenhall2020nerf}.
An extension to reflectance fields (with \cite{bi2020neural}) is discussed in Sec.~\ref{sec:reflectance}.

As shown in Fig.~\ref{fig:arch}, our method is composed of four learned components:$\,\FuncDens, \FuncTex, \FuncUV$ and $\FuncInv$. 
Unlike NeRF, which uses a single MLP, \netex{} uses a disentangled neural representation consisting of three sub-networks, which encode scene geometry ($\FuncDens$), a texture mapping function ($\FuncUV$), and a 2D texture ($\FuncTex$) respectively (Sec.~\ref{sec:representation}).
In addition, we use an inverse texture mapping network ($\FuncInv$) to ensure that the discovered texture space reasonably explains the scene surfaces (Sec.~\ref{sec:inverse}).
We introduce a cycle mapping loss to regularize the texture mapping and inverse mapping networks, and use a rendering loss to train our neural model end-to-end to regress realistic images (Sec.~\ref{sec:loss}).

\subsection{Disentangled neural scene representation}
\label{sec:representation}



\boldstart{Volume rendering.} Volume rendering requires volume density $\Dens$ and radiance $\Color$ at all 3D locations in a scene. 
A pixel's radiance value (RGB color) $\PixelColor$ is computed by marching a ray from the pixel and aggregating the radiance values $\Color_i$ of multiple shading points on the ray, as expressed by:
\begingroup
\setlength{\abovedisplayskip}{0.5em}
\setlength{\belowdisplayskip}{0.3em}
\setlength{\abovedisplayshortskip}{0pt}
\setlength{\belowdisplayshortskip}{0pt}
\addtolength{\jot}{-5pt}
\begin{align}
    \PixelColor &=  \sum_i\Trans_i (1-\exp (-\Dens_i\Step_i)) \Color_i, \label{eqn:raymarching} \\
    \Trans_i &= \exp (-\sum_{j=1}^{i-1} \Dens_j \Step_j ), \label{eqn:trans}
\end{align}
\endgroup
where $i = 1,...,N$ denotes the index of a shading point on the ray,  $\Step_i$ represents the distance between two consecutive points, $\Trans_i$ is known as the transmittance, and $\Color_i$ and $\Dens_i$ are the volume density (extinction coefficient) and radiance at shading point $i$.  The above ray marching process is derived as a discretization of a continuous volume rendering integral; for more details, please see previous work \cite{max1995optical}.

\boldstartspace{Radiance field.} In the context of view synthesis,
a general volume scene representation can be seen as a 5D function (i.e. a radiance field, as referred to by \cite{mildenhall2020nerf}):
\begin{equation}
    \Func_{\Dens,\Color}:(\Pos, \Dir) \rightarrow (\Dens, \Color),
    \label{eqn:NeRF}
\end{equation} 
which outputs
volume density and radiance $(\Dens, \Color)$ given a 3D location $\Pos=(x, y, z)$ and viewing direction $\Dir=(\theta, \phi)$. 
NeRF \cite{mildenhall2020nerf} proposes to use a single MLP network to represent $\Func_{\Dens,\Color}$ as a neural radiance field and achieves photo-realistic rendering results.
Their single network encapsulates the entire scene geometry and appearance as a whole; however, this ``bakes'' the scene content into the trained network, and does not allow for any applications (e.g., appearance editing) beyond pure view synthesis.

\boldstartspace{Disentangling $\Func_{\Dens,\Color}$.} In contrast, we propose explicitly decomposing the radiance field $\Func_{\Dens,\Color}$ into two components, $\Func_{\Dens}$ and $\Func_{\Color}$, modeling geometry and appearance, respectively:
\begin{equation}
    \Func_{\Dens}:\Pos \rightarrow \Dens, \quad\quad  \Func_{\Color}:(\Pos, \Dir) \rightarrow \Color.
    \label{eqn:GeoApp}
\end{equation} 
In particular, $\Func_{\Dens}$ regresses volume density (i.e., scene geometry), and $\Func_{\Color}$ regresses radiance (i.e., scene appearance). 
We model them as two independent networks.

\boldstartspace{Texture mapping.} 
We further propose to model scene appearance in a 2D texture space that explains the object's 2D surface appearance.
We explicitly map a 3D point $\Pos=(x,y,z)$ in a volume onto a 2D UV coordinate $\UV=(u,v)$ in a texture, and regress the radiance in the texture space given 2D UV coordinates and a viewing direction $(\UV, \Dir)$.
We describe the 3D-to-2D mapping as a texture mapping function $\FuncUV$ and the radiance regression as a texture function $\FuncTex$:
\begin{equation}
    \FuncUV:\Pos \rightarrow \UV, \quad\quad  \FuncTex:(\UV, \Dir) \rightarrow \Color.
    \label{eqn:texture}
\end{equation} 
Our appearance function $\Func_{\Color}$ is thus a composition of the two functions:
\begin{equation}
   \Func_{\Color}(\Pos, \Dir) = \FuncTex(\FuncUV(\Pos), \Dir).
\end{equation} 

\boldstartspace{Neural representation.}
In summary, our full radiance field is a composition of three functions: a geometry function $\Func_{\Dens}$, a texture mapping function $\FuncUV$, and a texture function $\FuncTex$, given by:
\begin{equation}
    (\Dens, \Color) = \Func_{\Dens,\Color}(\Pos, \Dir) =(\FuncDens(\Pos), \FuncTex(\FuncUV(\Pos), \Dir)).
    \label{eqn:neuralrep}
 \end{equation} 
We use three separate MLP networks for $\Func_{\Dens}$, $\FuncUV$ and $\FuncTex$. Unlike the black-box NeRF network, our representation has disentangled geometry and appearance modules, and models appearance in a 2D texture space. 

\subsection{Texture space and inverse texture mapping}
\label{sec:inverse}
As described in Eqn.~\ref{eqn:texture}, our texture space is parameterized by a 2D UV coordinate $\UV=(u,v)$.
While any continuous 2D topology can be used for the UV space in our network, 
we use a 2D unit sphere for most results, where $\UV$ is interpreted as a point on the unit sphere. 

Directly training the representation networks ($\FuncDens$, $\FuncUV$, $\FuncTex$) with pure rendering supervision often leads to a highly distorted texture space and degenerate cases where multiple points map to the same UV coordinate, which is undesirable.
The ideal goal is instead to uniformly map the 2D surface onto the texture space and occupy the entire texture space.
To achieve this, we propose to jointly train an ``inverse'' texture mapping network $\FuncInv$ that maps a 2D UV coordinate $\UV$ on the texture to a 3D point $\Pos$ in the volume:
\begin{equation}
    \FuncInv:\UV \rightarrow \Pos.
    \label{eqn:inverse}
\end{equation} 

$\FuncInv$ projects the 2D texture space onto a 2D manifold (in 3D space).
This inverse texture mapping allows us to reason about the 2D surface of the scene (corresponding to the inferred texture) and regularize the texture mapping process.
We leverage our texture mapping and inverse mapping networks to build a cycle mapping (a one-to-one correspondence) between the 2D object surface and the texture space, leading to high-quality texture mapping.

\subsection{Training neural texture mapping}
\label{sec:loss}
We train our full network, consisting of $\FuncDens$, $\FuncTex$, $\FuncUV$, and $\FuncInv$, from end to end, to simultaneously achieve surface discovery, space mapping, and scene geometry and appearance inference.

\boldstartspace{Rendering loss.}
We directly use the ground truth pixel radiance value $\PixelColor_\text{gt}$ in the captured images to supervise our rendered pixel radiance value $\PixelColor$ from ray marching (Eqn.~\ref{eqn:raymarching}).
The rendering loss for a pixel ray is given by:
 \begin{equation}
    L_\text{render} = \|\PixelColor_\text{gt} - \PixelColor\|_2^2.
\end{equation}
This the main source of supervision in our system.

\begin{figure}[t]
    \centering
    \includegraphics[width=\linewidth]{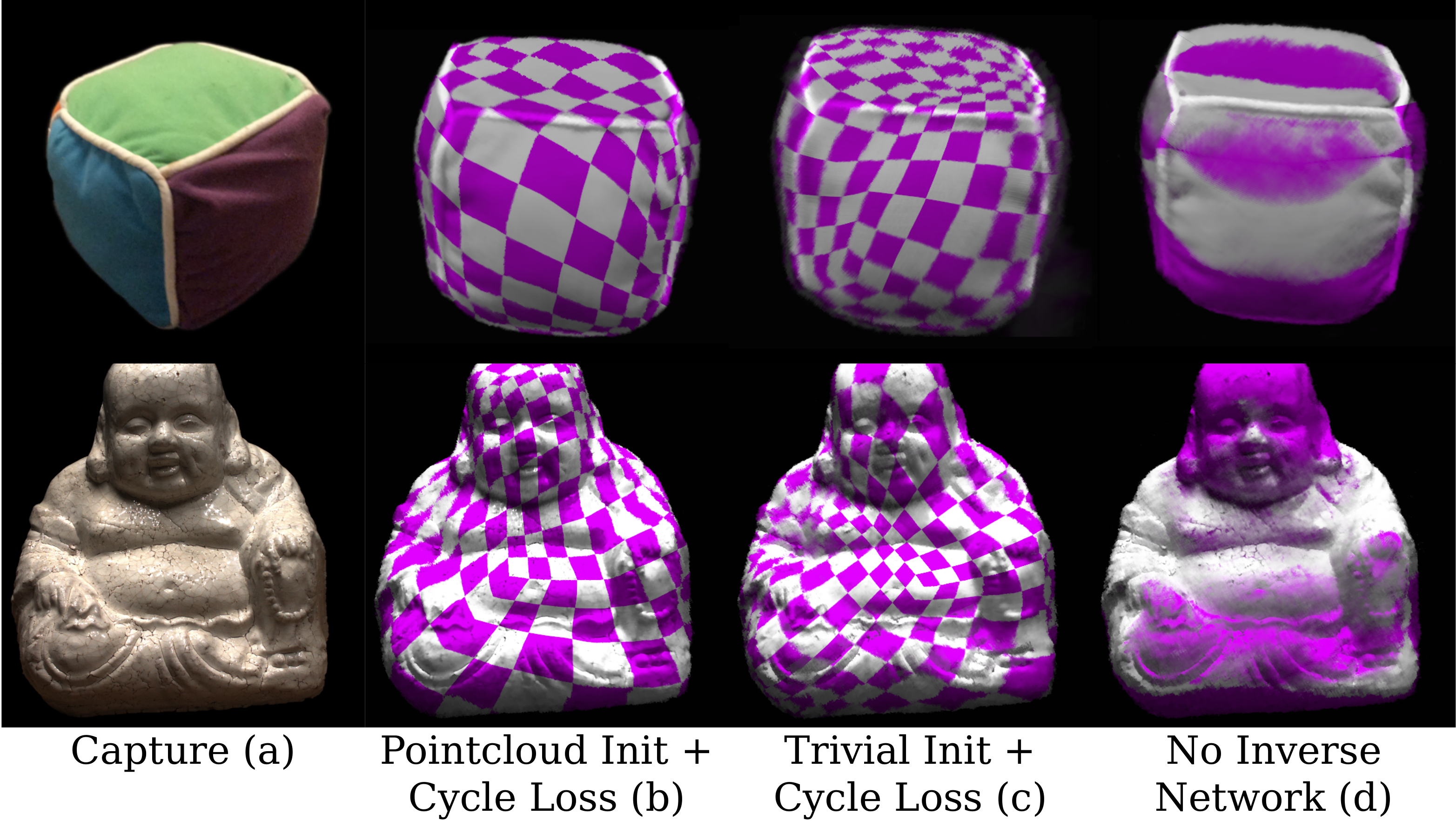}
    \caption{A checkerboard texture applied to scenes (a). When trained with or without initialization using coarse point cloud (b,c), the learned texture space is relatively uniform compared to trained without $F_{uv}^{-1}$ and cycle loss (d). 
 }
    \label{fig:inverse}
\end{figure}

\boldstartspace{Cycle loss.}
Given any sampled shading point $\Pos_i$ on a ray in ray marching,  
our texture mapping network finds its UV $\UV_i$ in texture space for radiance regression. 
We use the inverse mapping network to map this UV $\UV_i$ back to the 3D space:
\begin{equation}
    \Pos'_i= \FuncInv(\FuncUV(\Pos_i)).
\end{equation}
We propose to minimize the difference between $\Pos'_i$ and $\Pos_i$ to enforce a cycle mapping between the texture and world spaces (and
force $\FuncInv$ to learn the inverse of $\FuncUV$).

However, it is unnecessary and unreasonable to enforce a cycle mapping at any 3D point. 
We only expect a correspondence between the texture space and points on the 2D \emph{surface} of the scene; enforcing the cycle mapping in the empty space away from the surface is meaningless. 
We expect 3D points near the scene surface to have high contributions to the radiance.
Therefore, we leverage the radiance contribution weights per shading point to weigh our cycle loss.
Specifically, we consider the weight:
\begin{equation}
    \BlendW_i = \Trans_i (1-\exp (-\Dens_i\Step_i)),
\end{equation}
which determines the contribution to the final pixel color for each shading point $i$ in the ray marching equation~\ref{eqn:raymarching}. Equation~\ref{eqn:raymarching}
 can be simply written as $\PixelColor =  \sum_i \BlendW_i \Color_i$.
 This contribution weight $\BlendW_i$ naturally expresses how close a point is to the surface, and has been previously used for depth inference \cite{mildenhall2020nerf}. 
 Our cycle loss for a single ray is given by:
 \begin{equation}
    L_\text{cycle} = \sum_i \BlendW_i \|\FuncInv(\FuncUV(\Pos_i))-\Pos_i\|_2^2.
    \label{eqn:cycleloss}
\end{equation}

\boldstartspace{Mask loss.}
We also additionally provide a loss to supervise a foreground-background mask.
Basically, the transmittance (Eqn.~\ref{eqn:trans}) of the the last shading point $\Trans_N$ on a pixel ray indicates if the pixel is part of the background. 
We use the ground truth mask $\PixelMask_\text{gt}$ per pixel to supervise this by
\begin{equation}
    L_\text{mask} = \|\PixelMask_\text{gt} - (1-\Trans_N)\|_2^2.
\end{equation}
We found this mask loss is necessary when viewpoints do not cover the object entirely. 
In such cases, the network can use the volume density to darken (when the background is black) renderings and fake some shading effects that should be in the texture.
When the view coverage is dense enough around an object, this mask loss is often optional.

\boldstartspace{Full loss.}
Our full loss function $L$ during training is:
\begin{equation}
    L = L_\text{render} + a_1L_\text{cycle} + a_2L_\text{mask}.
    \label{eqn:loss}
\end{equation}
We use $a_1=1$ for all our scenes in our experiments.
We use $a_2=1$ for most scenes, except for those that already have good view coverage, where we remove the mask loss by setting $a_2=0$.

\Comment{
\subsection{Extension to reflectance fields}
\label{sec:reflectance}

\KS{wondering if this section should be moved to results/applications.}
\netex{} can be incorporated into different volume rendering pipelines.
We have introduced applying it to NeRF for view synthesis in Secs.~\ref{sec:representation} to \ref{sec:loss}.
We now briefly discuss combining it with the recent Neural Reflectance Fields \cite{bi2020neural} work that reconstructs BRDFs in volume rendering from flash images.

Instead of directly outputting radiance $\Color$ at each shading point, this work \cite{bi2020deep} regresses normal $\Normal$ and reflectance parameters $\Refl$ at each shading point, and introduces a reflectance-aware volume rendering that computes radiance from these shading properties under given viewing and lighting condition.
We correspondingly modify our geometry network $\FuncDens$ to jointly regress volume density and normal, and change the texture regression network $\FuncTex$ to regress the reflectance parameters in the texture space. 
Our central texture mapping and inverse mapping networks remain the same for this case.
The modified network naturally provides the required volume properties in the reflectance-aware volume rendering process. 
We show that our neural texture mapping can enable high-quality BRDF texture extraction in this setting.
}

\section{Implementation Details}
\subsection{Network details}
All four sub-networks, $\FuncDens$, $\FuncTex$, $\FuncUV$, and $\FuncInv$, are designed as MLP networks.
We use unit vectors to represent viewing direction $\Dir$ and UV coordinate $\UV$ (for spherical UV). 
As proposed by NeRF, we use positional encoding to infer high-frequency geometry and appearance details. 
In particular, we apply positional encoding for our geometry network $\FuncDens$ and texture network $\FuncTex$ on all their input components including $\Pos$, $\UV$ and $\Dir$. 
On the other hand, since the texture mapping is expected to be smooth and uniform, we do not apply positional encoding on the two mapping networks. Please refer to the supplemental materials for the detailed architecture of our networks.

\begin{figure*}[t]
    \centering
    \includegraphics[width=0.8\linewidth]{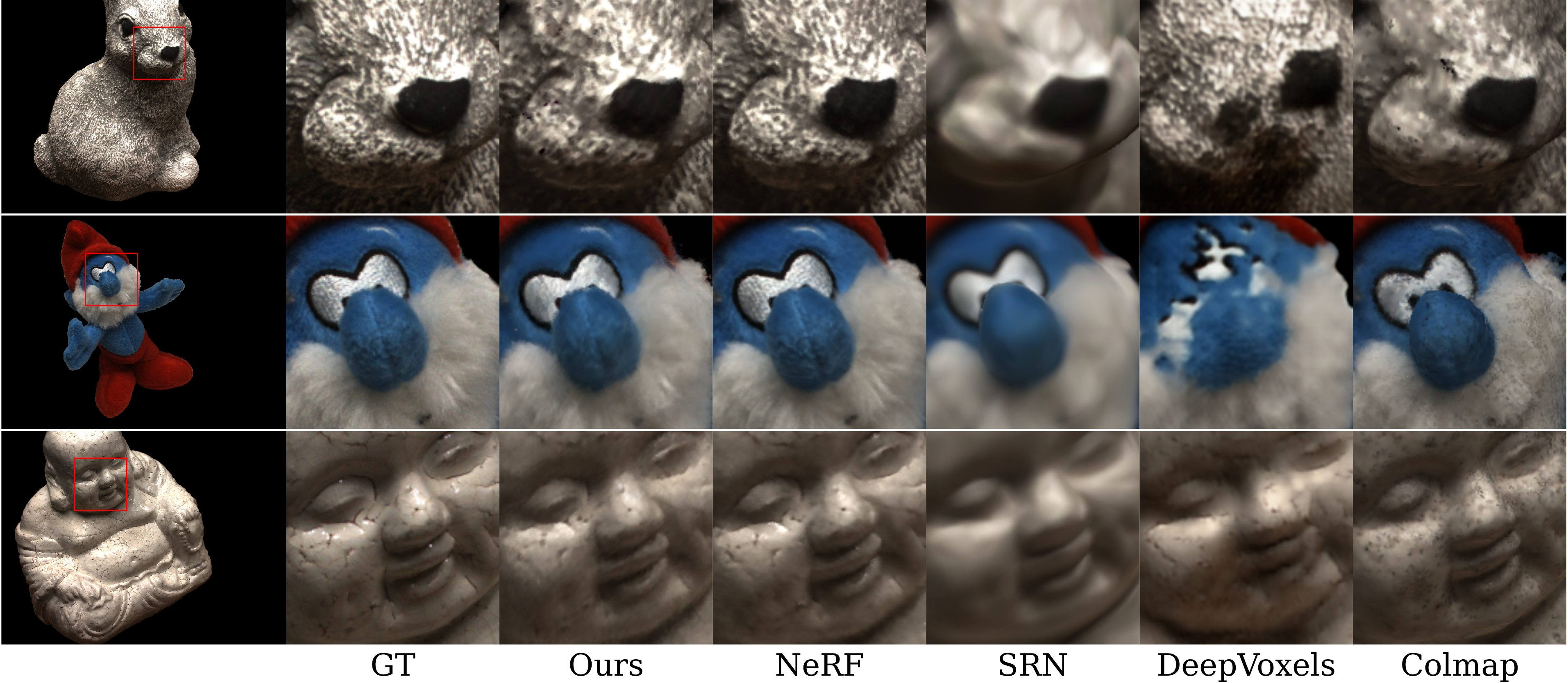}
    \caption{Comparisons on DTU scenes. Note how visually close our method is to the state-of-the-art, while enabling editing.}
    \label{fig:dtu}
\end{figure*}

\subsection{Training details}
\label{sec:traindetails}
Before training, we normalize the scene space to the unit box. 
When generating rays, we sample shading points on each pixel ray inside the box.
For all our experiments, we use stratified sampling (uniform sampling with local jittering) to sample 256 point on each ray for ray marching. 
For each iteration, we randomly select a batch size of 600 to 800 pixels (depending on GPU memory usage) from an input image; we take 2/3 pixels from the foreground and 1/3 pixels from the background. 

Our inverse mapping network $\FuncInv$ maps the 2D UV space to a 3D surface, which is functionally similar to AtlasNet \cite{groueix2018papier} and can be trained as such, if geometry is available. 
We thus initialize $\FuncInv$ with a point cloud from COLMAP \cite{schoenberger2016mvs} using a Chamfer loss.
However, since the MVS point cloud is often very noisy, this Chamfer loss is only used during this initialization phase.
We find this initialization facilitates training, though our network still works without it for most cases (see Fig.~\ref{fig:inverse}).
Usually, this AtlasNet-style initialization is very sensitive to the MVS reconstruction noise and leads to a highly non-uniform mapping surface.
However, we find that our final inverse mapping network can output a much smoother surface as shown in Fig.~\ref{fig:atlas}, after jointly training with our rendering and cycle losses.

Specifically, we initially train our method using a Chamfer loss together with a rendering loss for 50,000 iterations. 
Then, we remove the Chamfer loss and train with our full loss (Eqn.~\ref{eqn:loss}) until convergence, after around 500,000 iterations. 
Finally, we fine-tune our texture network $\FuncTex$ until convergence, freezing the other networks ($\FuncDens$, $\FuncUV$ and $\FuncInv$), which is useful to get better texture details. 
The whole process takes 2-3 days on a single RTX 2080Ti GPU.


\section{Results}
\Comment{
\begin{figure*}
    \centering
    \includegraphics[width=\textwidth]{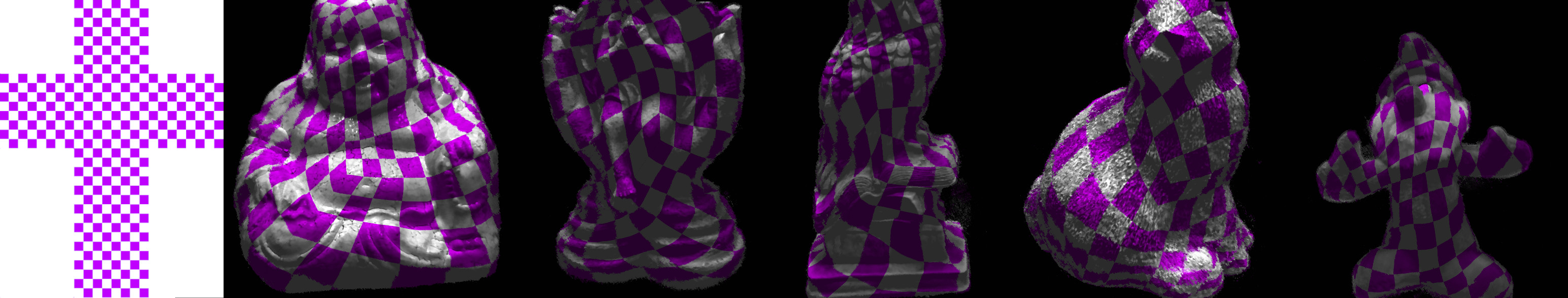}
    \caption{Apply checkerboard texture (left) to the DTU objects. Our approach discovers a uniform texture space.}
    \label{fig:checkerboard}
\end{figure*}
}


\begin{table}[t]
\centering
\begin{tabular}{l|ll}
Method & PSNR & SSIM \\ \hline
SRN \cite{sitzmann2019scene} & 26.05 & 0.837 \\
DeepVoxels \cite{sitzmann2019deepvoxels} & 20.85 & 0.702 \\
Colmap \cite{schoenberger2016mvs} & 24.63 & 0.865 \\
NeRF\cite{mildenhall2020nerf} & \textbf{30.73} & \textbf{0.938} \\
Ours & \textbf{28.23} & \textbf{0.894}
\end{tabular}
\caption{Average PSNR/SSIM for novel view synthesis on 4 held-out views on 5 DTU scenes. See supplementary for full table.}
\label{tab:number}
\end{table}

\begin{figure*}[t]
    \centering
    \includegraphics[width=1\linewidth]{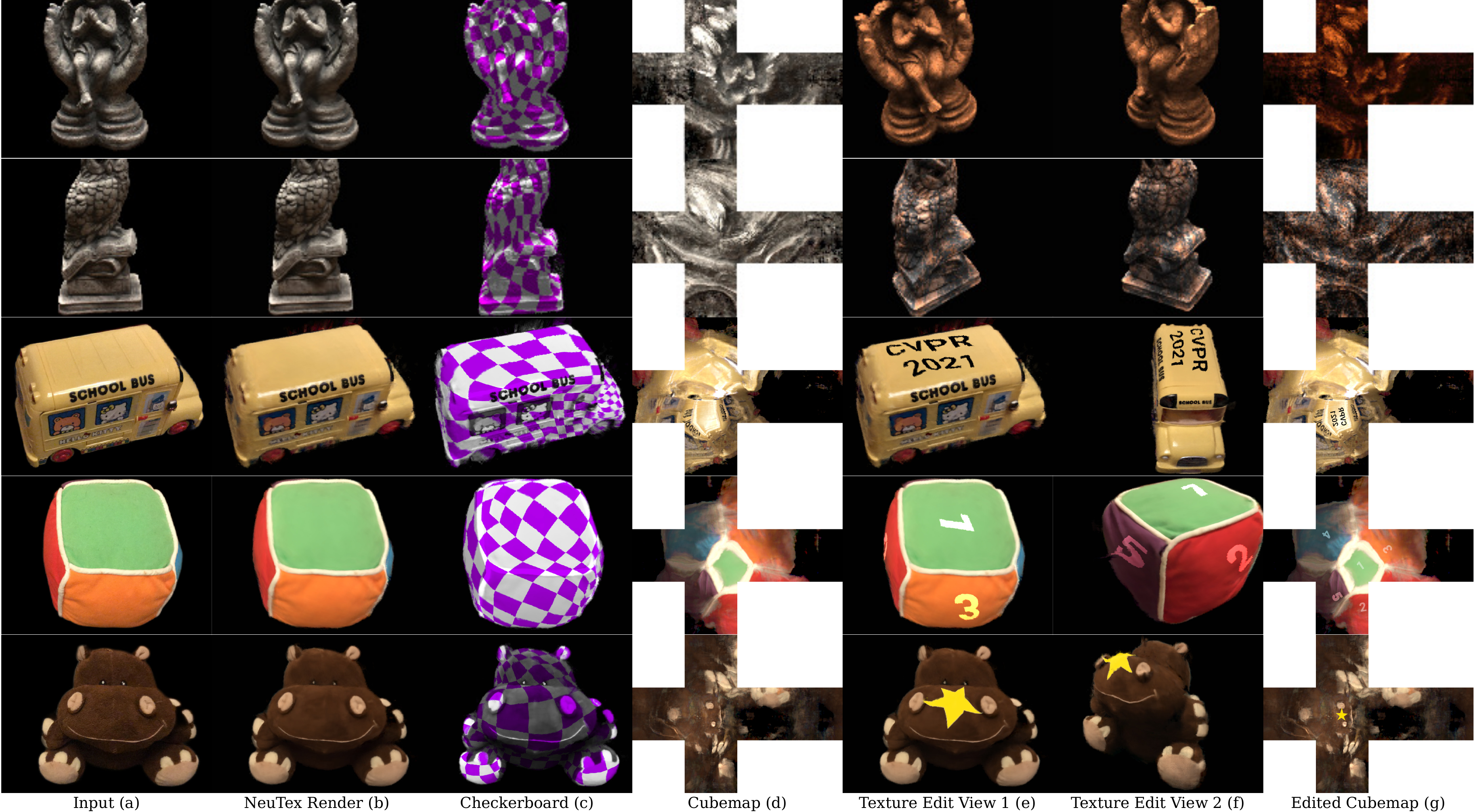}
    \caption{Texture editing on DTU (rows 1-2) and our (rows 3-5) scenes. Since our neural texture ($\FuncTex$) depends on a view direction, we show the cubemap texture (d) with pixelwise maximum values across views. Each texture is edited by multiplying a new specified texture; the resulting texture is shown in (g). The images rendered with the edited textures are shown from two different views in (e) and (f).}
    \label{fig:results}
\end{figure*}

We now show experimental results of our method and comparisons against previous methods on real scenes.

\subsection{Configuration}

\begin{figure*}[t]
    \centering
    \includegraphics[width=\linewidth]{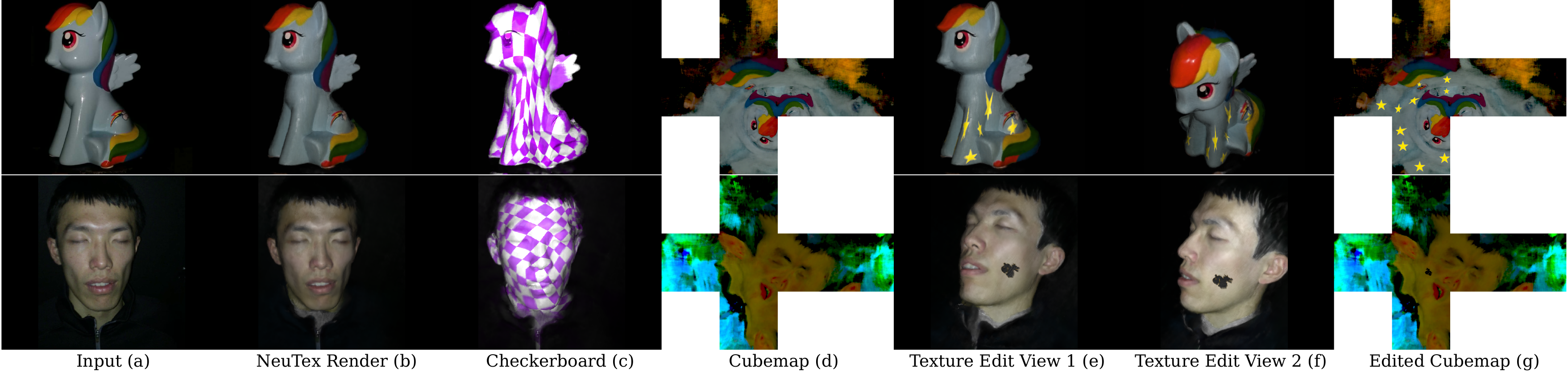}
    \caption{\netex{} in reflectance fields setting. We edit captured diffuse albedo (d) as shown in (g) to produce results shown in (e,f). }
    \label{fig:refresults}
\end{figure*}

We demonstrate our method on real scenes from different sources, including five scenes from the DTU dataset \cite{aanaes2016large} (Fig.~\ref{fig:teaser}, \ref{fig:dtu}, \ref{fig:results}), two scenes from Neural Reflectance Fields \cite{bi2020neural} obtained from the authors (Fig.~\ref{fig:refresults}), and three scenes captured by ourselves (Fig.~\ref{fig:results}). 
Each DTU scene contains either 49 or 64 input images from multiple viewpoints. 
Each scene from \cite{bi2020neural} contains about 300 images.
Our own scenes each contain about 100 images.
For our own data, we capture the images using a held-hanld cellphone and use the structure from motion implementation in COLMAP \cite{schoenberger2016sfm} for camera calibration. 
For other scenes, we directly use the provided camera calibration in the dataset. 
Since our method focuses on the capture and surface discovery of objects, we require the input images to have a clean, easily segmentable background. 
We use U2Net \cite{Qin_2020_PR} to automatically compute masks for our own scenes. 
For the DTU scenes, we use the background masks provided by \cite{yariv2020multiview}.
The images from \cite{bi2020neural} are captured under a single flash light, which already have very dark background; thus we do not apply additional masks for these images. 


\subsection{View synthesis results on DTU scenes}
\label{sec:viewsyncomp}
We now evaluate and compare our view synthesis results on five DTU scenes. 
In particular, we compare with NeRF \cite{mildenhall2020nerf}, two previous neural rendering methods, SRN \cite{sitzmann2019scene} and DeepVoxels \cite{sitzmann2019deepvoxels}, and one classical mesh reconstruction method COLMAP \cite{schoenberger2016mvs}.
We use the released code from their authors to generate the results for all the comparison methods.
For COLMAP, we skip the structure from motion, since we already have the provided camera calibration from the dataset.
We hold-out 4 random views as testing views from the original 49 or 64 input views and run all methods on the remaining images for reconstruction. 

We show qualitative visual comparisons on zoomed-in crops of testing images of two DTU scenes in Fig.~\ref{fig:dtu} (the other scenes are shown in supplementary materials), and quantitative comparison results of the averaged PSNRs and SSIMs on the testing images across five scenes in Tab.~\ref{tab:number}. 
Our method achieves high-quality view synthesis results as reflected by our rendered images being close to the ground truth and also our high PSNRs and SSIMs. 
Note that \netex{} enables automatic texture mapping that none of the other comparison methods can do. Even a traditional mesh-based method like COLMAP \cite{schoenberger2016mvs} needs additional techniques or tools to unwrap its surface for texture mapping, whereas our method unwraps the surface into a texture while doing reconstruction in a unsupervised way. 
To achieve this challenging task, \netex{} is designed in a more constrained way than NeRF.
As a result, our rendering quality is quantitatively slightly worse than NeRF. 
Nonetheless, as shown in Fig.~\ref{fig:dtu}, our rendered results are realistic, reproduce many high-frequency details and qualitatively look very close to NeRF's results. 

In fact, our results are significantly better than all other comparison methods, including both mesh-based reconstruction \cite{schoenberger2016mvs} and previous neural rendering methods \cite{sitzmann2019deepvoxels,sitzmann2019scene} in both qualitative and quantitative comparisons.
In particular, COLMAP \cite{schoenberger2016sfm} can reconstruct reasonable shapes, but it cannot recover accurate texture details and intrinsically lacks view-dependent appearance effects (due to the Lambertian materials assumption).
DeepVoxels \cite{sitzmann2019deepvoxels} leverages a non-physically-based module for geometry and occlusion inference. 
While this works well on scenes that have hundreds of input images, it does not work well on DTU scenes that have only about 40 to 60 images, leading to incorrect shapes and serious artifacts in their results. 
SRN \cite{sitzmann2019scene}, on the other hand, can reproduce reasonable shape in the rendering; however it cannot generate high-frequency appearance details like our method. 
Our approach is based on physically-based volume rendering, which models scene geometry and appearance accurately, leading to photo-realistic rendering results.
More importantly, our approach achieves texture mapping and enables surface appearance editing in a 2D texture space that it automatically discovered, which cannot be done by NeRF nor any other previous neural rendering approaches.

\subsection{Texture mapping and appearance editing}
\label{sec:mapresults}
We now demonstrate our unique results on texture mapping and texture-space appearance editing that previous neural rendering approaches cannot achieve.
Figure~\ref{fig:results} shows such results on diverse real objects of DTU scenes and our own captured scenes.
Our method can synthesize realistic view synthesis results (Fig.~\ref{fig:results}.b) that are very close to the ground truth.
In addition, our method successfully unwraps the object surface into a reasonable texture (Fig.~\ref{fig:results}.d);
the discovered texture space meaningfully expresses the 2D surface and distributes uniformly, as illustrated by the checkerboard rendering shown in Fig.~\ref{fig:results}.c.

\boldstartspace{Texture editing.} Our high-quality texture mapping enables flexible appearance editing applications as shown in Fig.~\ref{fig:results}.e-g.
In these examples, we show that we can use a specified full texture map to modulate the original texture, which entirely changes the object appearance.
For example, the object in the 1st row is interestingly changed from a stone-like object to a wooden one.
We also demonstrate that we can locally modify the texture to add certain patterns on the object surface, such as the CVPR logo, the numbers, and the star in the last three rows.
Note that, all our appearance editing is directly done in the texture space, which changes the essential surface appearance and naturally appears consistent across multiple view points, as shown in the rendered images from two different views in Fig.~\ref{fig:results}.e and f.
Our \netex{} successfully disentangles the geometry and appearance of real objects and model the surface appearance in a meaningful texture space that explains the surface.

\begin{figure}[t]
    \centering
\includegraphics[width=0.9\linewidth]{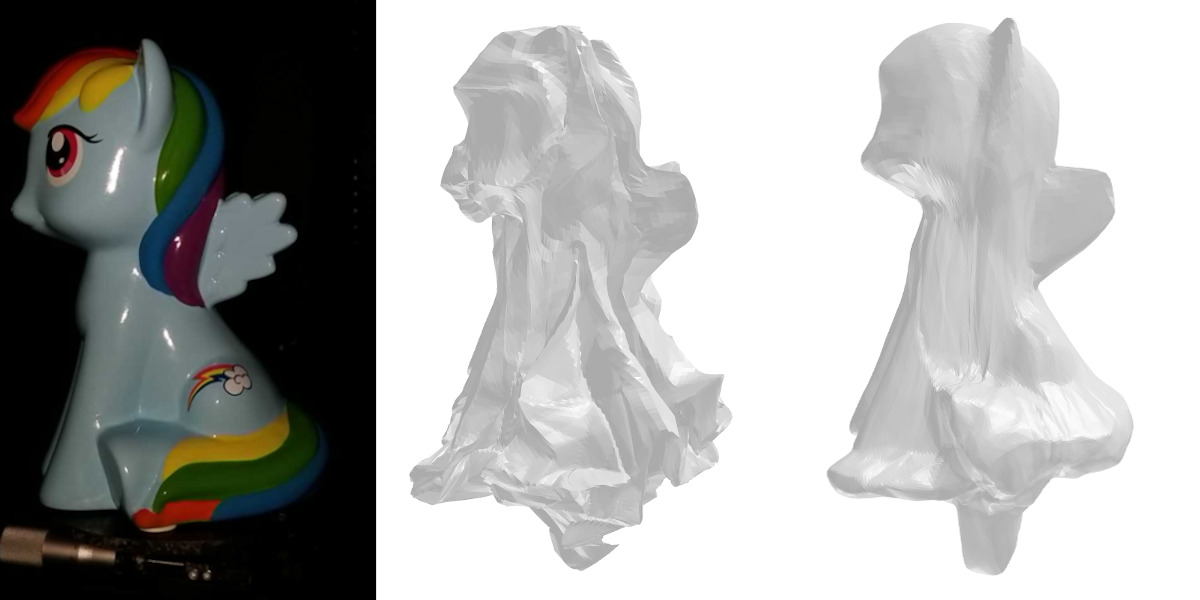}
    \caption{Our parametric surface ($F_{uv}^{-1}$) is strongly affected by noise in the COLMAP point cloud when trained with a Chamfer loss (center). Finetuning with our cycle and rendering losses without point cloud supervision (right) gives a smoother surface. }
    \label{fig:atlas}
\end{figure}

\boldstartspace{Inverse mapping.}
We further demonstrate the role of our inverse mapping network $\FuncInv$ in discovering this reasonable texture space.
When we remove $\FuncInv$ from our network and train the system with only the rendering loss, the result always leads to a degenerate texture mapping, where large regions of 3D points are mapped to the same UV coordinates, as illustrated by the checkerboard-texture rendering shown in Fig.~\ref{fig:inverse}. In contrast, our full network with the cycle loss generally discovers a uniform space.
As described in Sec.~\ref{sec:traindetails}, we initialize our inverse network using a point cloud with a Chamfer loss; this is done mainly to help the network converges quickly to a reasonable stage.
In Fig.~\ref{fig:inverse}, we also show that our inverse network can still function well even without the point cloud initialization, using only the cycle loss. 
Note that, our initial point clouds come from an MVS reconstruction \cite{schoenberger2016mvs}, which contains a lot of noise, leading to a noisy surface out of $\FuncInv$ as shown in Fig.~\ref{fig:atlas}. To prevent a degradation in the surface texture quality, we remove the supervision on this initial point cloud after initialization. 
Instead, our cycle loss can continue improving the noisy initialization and let the inverse mapping network $\FuncInv$ discover a smooth surface as shown in Fig.~\ref{fig:atlas}.

\subsection{Extension to reflectance fields}
\label{sec:reflectance}
\netex{} can be incorporated into different volume rendering pipelines.
We now discuss combining it with the recent Neural Reflectance Fields \cite{bi2020deep,bi2020neural} that reconstructs BRDFs in volume rendering from flash images. 

Instead of directly outputting radiance $\Color$, \cite{bi2020neural} regresses normal $\Normal$ and reflectance parameters $\Refl$ at each shading point, and introduces a reflectance-aware volume rendering that computes radiance from these shading properties under given viewing and lighting condition.
We correspondingly modify our geometry network $\FuncDens$ to jointly regress volume density and normal, and change the texture regression network $\FuncTex$ to regress the reflectance parameters in the texture space. 
Our central texture mapping and inverse mapping networks remain the same for this case.
The modified network naturally provides the required volume properties in the reflectance-aware volume rendering process. 

We show that our neural texture mapping can enable high-quality BRDF texture extraction in this setting in Fig.~\ref{fig:refresults} on the two scenes from \cite{bi2020neural}.
Our approach achieves realistic rendering (Fig.~\ref{fig:refresults}.b) that reproduces the original appearance, discovers a reasonably uniform texture space (Fig.~\ref{fig:refresults}.c), successfully unwraps the surface BRDFs into this space (as shown by the albedo maps in Fig.~\ref{fig:refresults}.d), and enables realistic rendering with BRDF editing in the texture space (Fig.~\ref{fig:refresults}.e-g).
These results demonstrate the generality of our neural texture mapping framework and inspire potential future applications of our technique on other neural rendering tasks.

\section{Conclusion}

We have presented a novel approach that enables texture mapping in neural volumetric rendering.
We introduce a novel disentangled neural scene representation that models geometry as a 3D volume and models appearance as a 2D texture in a automatically discovered texture space.
We propose to jointly train a 3D-to-2D texture mapping network and a 2D-to-3D inverse mapping network to achieve surface reasoning and texture space discovery, using a surface-aware cycle consistency loss.
As demonstrated, our approach can discover a reasonable texture space that meaningfully explains the object surface.
Our method enables flexible surface appearance editing applications for neural volumetric rendering.
\paragraph{Acknowledgement} This research was supported by gifts from Adobe, Kwai, and Qualcomm.


{\small
\bibliographystyle{ieee_fullname}
\bibliography{ms}

\begin{thebibliography}{10}\itemsep=-1pt

\bibitem{aanaes2016large}
Henrik Aan{\ae}s, Rasmus~Ramsb{\o}l Jensen, George Vogiatzis, Engin Tola, and
  Anders~Bjorholm Dahl.
\newblock Large-scale data for multiple-view stereopsis.
\newblock {\em International Journal of Computer Vision}, 120(2):153--168,
  2016.

\bibitem{achlioptas2018learning}
Panos Achlioptas, Olga Diamanti, Ioannis Mitliagkas, and Leonidas Guibas.
\newblock Learning representations and generative models for {3D} point clouds.
\newblock In {\em ICML}, pages 40--49, 2018.

\bibitem{bi2017patch}
Sai Bi, Nima~Khademi Kalantari, and Ravi Ramamoorthi.
\newblock Patch-based optimization for image-based texture mapping.
\newblock {\em ACM Transaction on Graphics}, 36(4):106--1, 2017.

\bibitem{bi2020neural}
Sai Bi, Zexiang Xu, Pratul Srinivasan, Ben Mildenhall, Kalyan Sunkavalli,
  Milo{\v{s}} Ha{\v{s}}an, Yannick Hold-Geoffroy, David Kriegman, and Ravi
  Ramamoorthi.
\newblock Neural reflectance fields for appearance acquisition.
\newblock {\em arXiv preprint arXiv:2008.03824}, 2020.

\bibitem{bi2020deep}
Sai Bi, Zexiang Xu, Kalyan Sunkavalli, Milo{\v{s}} Ha{\v{s}}an, Yannick
  Hold-Geoffroy, David Kriegman, and Ravi Ramamoorthi.
\newblock Deep reflectance volumes: Relightable reconstructions from multi-view
  photometric images.
\newblock {\em arXiv preprint arXiv:2007.09892}, 2020.

\bibitem{buehler2001unstructured}
Chris Buehler, Michael Bosse, Leonard McMillan, Steven Gortler, and Michael
  Cohen.
\newblock Unstructured lumigraph rendering.
\newblock In {\em SIGGRAPH}, pages 425--432. ACM, 2001.

\bibitem{chen2019point}
Rui Chen, Songfang Han, Jing Xu, and Hao Su.
\newblock Point-based multi-view stereo network.
\newblock In {\em Proceedings of the IEEE International Conference on Computer
  Vision}, pages 1538--1547, 2019.

\bibitem{chen2020neural}
Zhang Chen, Anpei Chen, Guli Zhang, Chengyuan Wang, Yu Ji, Kiriakos~N.
  Kutulakos, and Jingyi Yu.
\newblock A neural rendering framework for free-viewpoint relighting.
\newblock In {\em CVPR}, June 2020.

\bibitem{chen2018learning}
Zhiqin Chen and Hao Zhang.
\newblock Learning implicit fields for generative shape modeling.
\newblock {\em arXiv preprint arXiv:1812.02822}, 2018.

\bibitem{cheng2020deep}
Shuo Cheng, Zexiang Xu, Shilin Zhu, Zhuwen Li, Li~Erran Li, Ravi Ramamoorthi,
  and Hao Su.
\newblock Deep stereo using adaptive thin volume representation with
  uncertainty awareness.
\newblock In {\em Proceedings of the IEEE/CVF Conference on Computer Vision and
  Pattern Recognition}, pages 2524--2534, 2020.

\bibitem{debevec1998efficient}
Paul Debevec, Yizhou Yu, and George Borshukov.
\newblock Efficient view-dependent image-based rendering with projective
  texture-mapping.
\newblock In {\em Rendering Techniques’ 98}, pages 105--116. Springer, 1998.

\bibitem{furukawa2009accurate}
Yasutaka Furukawa and Jean Ponce.
\newblock Accurate, dense, and robust multiview stereopsis.
\newblock {\em IEEE transactions on pattern analysis and machine intelligence},
  32(8):1362--1376, 2009.

\bibitem{goel2020shape}
Shubham Goel, Angjoo Kanazawa, and Jitendra Malik.
\newblock Shape and viewpoint without keypoints.
\newblock {\em arXiv preprint arXiv:2007.10982}, 2020.

\bibitem{groueix2018papier}
Thibault Groueix, Matthew Fisher, Vladimir~G Kim, Bryan~C Russell, and Mathieu
  Aubry.
\newblock A papier-m{\^a}ch{\'e} approach to learning 3d surface generation.
\newblock In {\em Proceedings of the IEEE conference on computer vision and
  pattern recognition}, pages 216--224, 2018.

\bibitem{hedman2018deep}
Peter Hedman, Julien Philip, True Price, Jan-Michael Frahm, George Drettakis,
  and Gabriel Brostow.
\newblock Deep blending for free-viewpoint image-based rendering.
\newblock {\em ACM Transactions on Graphics (TOG)}, 37(6):1--15, 2018.

\bibitem{henzler2020learning}
Philipp Henzler, Niloy~J Mitra, and Tobias Ritschel.
\newblock Learning a neural 3d texture space from 2d exemplars.
\newblock In {\em Proceedings of the IEEE/CVF Conference on Computer Vision and
  Pattern Recognition}, pages 8356--8364, 2020.

\bibitem{huang2018deepmvs}
Po-Han Huang, Kevin Matzen, Johannes Kopf, Narendra Ahuja, and Jia-Bin Huang.
\newblock {DeepMVS}: Learning multi-view stereopsis.
\newblock In {\em CVPR}, pages 2821--2830, 2018.

\bibitem{ji2017surfacenet}
Mengqi Ji, Juergen Gall, Haitian Zheng, Yebin Liu, and Lu Fang.
\newblock {SurfaceNet}: An end-to-end {3D} neural network for multiview
  stereopsis.
\newblock In {\em ICCV}, pages 2307--2315, 2017.

\bibitem{kanazawa2018learning}
Angjoo Kanazawa, Shubham Tulsiani, Alexei~A Efros, and Jitendra Malik.
\newblock Learning category-specific mesh reconstruction from image
  collections.
\newblock In {\em Proceedings of the European Conference on Computer Vision
  (ECCV)}, pages 371--386, 2018.

\bibitem{kazhdan2006poisson}
Michael Kazhdan, Matthew Bolitho, and Hugues Hoppe.
\newblock Poisson surface reconstruction.
\newblock In {\em Proceedings of the fourth Eurographics symposium on Geometry
  processing}, volume~7, 2006.

\bibitem{kutulakos2000theory}
Kiriakos~N Kutulakos and Steven~M Seitz.
\newblock A theory of shape by space carving.
\newblock {\em ICCV}, 38(3):199--218, 2000.

\bibitem{liu2015learning}
Fayao Liu, Chunhua Shen, Guosheng Lin, and Ian Reid.
\newblock Learning depth from single monocular images using deep convolutional
  neural fields.
\newblock {\em IEEE transactions on pattern analysis and machine intelligence},
  38(10):2024--2039, 2015.

\bibitem{liu2020neural}
Lingjie Liu, Jiatao Gu, Kyaw Zaw~Lin, Tat-Seng Chua, and Christian Theobalt.
\newblock Neural sparse voxel fields.
\newblock {\em Advances in Neural Information Processing Systems}, 33, 2020.

\bibitem{lombardi2019neural}
Stephen Lombardi, Tomas Simon, Jason Saragih, Gabriel Schwartz, Andreas
  Lehrmann, and Yaser Sheikh.
\newblock Neural volumes: Learning dynamic renderable volumes from images.
\newblock {\em arXiv preprint arXiv:1906.07751}, 2019.

\bibitem{lorensen1987marching}
William~E Lorensen and Harvey~E Cline.
\newblock Marching cubes: A high resolution 3d surface construction algorithm.
\newblock {\em ACM siggraph computer graphics}, 21(4):163--169, 1987.

\bibitem{max1995optical}
Nelson Max.
\newblock Optical models for direct volume rendering.
\newblock {\em IEEE Transactions on Visualization and Computer Graphics},
  1(2):99--108, 1995.

\bibitem{mescheder2018occupancy}
Lars Mescheder, Michael Oechsle, Michael Niemeyer, Sebastian Nowozin, and
  Andreas Geiger.
\newblock Occupancy networks: Learning 3d reconstruction in function space.
\newblock {\em arXiv preprint arXiv:1812.03828}, 2018.

\bibitem{mildenhall2020nerf}
Ben Mildenhall, Pratul~P Srinivasan, Matthew Tancik, Jonathan~T Barron, Ravi
  Ramamoorthi, and Ren Ng.
\newblock Nerf: Representing scenes as neural radiance fields for view
  synthesis.
\newblock {\em arXiv preprint arXiv:2003.08934}, 2020.

\bibitem{niemeyer2020differentiable}
Michael Niemeyer, Lars Mescheder, Michael Oechsle, and Andreas Geiger.
\newblock Differentiable volumetric rendering: Learning implicit 3d
  representations without 3d supervision.
\newblock In {\em CVPR}, pages 3504--3515, 2020.

\bibitem{oechsle2019texture}
Michael Oechsle, Lars Mescheder, Michael Niemeyer, Thilo Strauss, and Andreas
  Geiger.
\newblock Texture fields: Learning texture representations in function space.
\newblock In {\em Proceedings of the IEEE International Conference on Computer
  Vision}, pages 4531--4540, 2019.

\bibitem{oechsle2020learning}
Michael Oechsle, Michael Niemeyer, Lars Mescheder, Thilo Strauss, and Andreas
  Geiger.
\newblock Learning implicit surface light fields.
\newblock {\em arXiv preprint arXiv:2003.12406}, 2020.

\bibitem{philip2019multi}
Julien Philip, Micha{\"e}l Gharbi, Tinghui Zhou, Alexei~A Efros, and George
  Drettakis.
\newblock Multi-view relighting using a geometry-aware network.
\newblock {\em ACM Transactions on Graphics}, 38(4):1--14, 2019.

\bibitem{poursaeed2020coupling}
Omid Poursaeed, Matthew Fisher, Noam Aigerman, and Vladimir~G Kim.
\newblock Coupling explicit and implicit surface representations for generative
  3d modeling.
\newblock {\em arXiv preprint arXiv:2007.10294}, 2, 2020.

\bibitem{qi2017pointnet}
Charles~R Qi, Hao Su, Kaichun Mo, and Leonidas~J Guibas.
\newblock Pointnet: Deep learning on point sets for 3d classification and
  segmentation.
\newblock In {\em Proceedings of the IEEE conference on computer vision and
  pattern recognition}, pages 652--660, 2017.

\bibitem{qi2016volumetric}
Charles~R Qi, Hao Su, Matthias Nie{\ss}ner, Angela Dai, Mengyuan Yan, and
  Leonidas~J Guibas.
\newblock Volumetric and multi-view cnns for object classification on 3d data.
\newblock In {\em Proceedings of the IEEE conference on computer vision and
  pattern recognition}, pages 5648--5656, 2016.

\bibitem{Qin_2020_PR}
Xuebin Qin, Zichen Zhang, Chenyang Huang, Masood Dehghan, Osmar Zaiane, and
  Martin Jagersand.
\newblock U2-net: Going deeper with nested u-structure for salient object
  detection.
\newblock {\em Pattern Recognition}, 106:107404, 2020.

\bibitem{saito2017photorealistic}
Shunsuke Saito, Lingyu Wei, Liwen Hu, Koki Nagano, and Hao Li.
\newblock Photorealistic facial texture inference using deep neural networks.
\newblock In {\em Proceedings of the IEEE Conference on Computer Vision and
  Pattern Recognition}, pages 5144--5153, 2017.

\bibitem{schoenberger2016sfm}
Johannes~Lutz Sch\"{o}nberger and Jan-Michael Frahm.
\newblock Structure-from-motion revisited.
\newblock In {\em Conference on Computer Vision and Pattern Recognition
  (CVPR)}, 2016.

\bibitem{schoenberger2016mvs}
Johannes~Lutz Sch\"{o}nberger, Enliang Zheng, Marc Pollefeys, and Jan-Michael
  Frahm.
\newblock Pixelwise view selection for unstructured multi-view stereo.
\newblock In {\em European Conference on Computer Vision (ECCV)}, 2016.

\bibitem{sitzmann2019deepvoxels}
Vincent Sitzmann, Justus Thies, Felix Heide, Matthias Nie{\ss}ner, Gordon
  Wetzstein, and Michael Zollhofer.
\newblock Deepvoxels: Learning persistent {3D} feature embeddings.
\newblock In {\em CVPR}, pages 2437--2446, 2019.

\bibitem{sitzmann2019scene}
Vincent Sitzmann, Michael Zollh{\"o}fer, and Gordon Wetzstein.
\newblock Scene representation networks: Continuous 3d-structure-aware neural
  scene representations.
\newblock In {\em Advances in Neural Information Processing Systems}, pages
  1119--1130, 2019.

\bibitem{tang2018ba}
Chengzhou Tang and Ping Tan.
\newblock Ba-net: Dense bundle adjustment network.
\newblock {\em arXiv preprint arXiv:1806.04807}, 2018.

\bibitem{thies2019deferred}
Justus Thies, Michael Zollh{\"o}fer, and Matthias Nie{\ss}ner.
\newblock Deferred neural rendering: Image synthesis using neural textures.
\newblock {\em ACM Transactions on Graphics (TOG)}, 38(4):1--12, 2019.

\bibitem{vijayanarasimhan2017sfm}
Sudheendra Vijayanarasimhan, Susanna Ricco, Cordelia Schmid, Rahul Sukthankar,
  and Katerina Fragkiadaki.
\newblock Sfm-net: Learning of structure and motion from video.
\newblock {\em arXiv preprint arXiv:1704.07804}, 2017.

\bibitem{wang2018mvpnet}
Jinglu Wang, Bo Sun, and Yan Lu.
\newblock Mvpnet: Multi-view point regression networks for {3D} object
  reconstruction from a single image.
\newblock {\em arXiv preprint arXiv:1811.09410}, 2018.

\bibitem{wang2018pixel2mesh}
Nanyang Wang, Yinda Zhang, Zhuwen Li, Yanwei Fu, Wei Liu, and Yu-Gang Jiang.
\newblock Pixel2mesh: Generating 3d mesh models from single rgb images.
\newblock In {\em Proceedings of the European Conference on Computer Vision
  (ECCV)}, pages 52--67, 2018.

\bibitem{wu20153d}
Zhirong Wu, Shuran Song, Aditya Khosla, Fisher Yu, Linguang Zhang, Xiaoou Tang,
  and Jianxiong Xiao.
\newblock 3d shapenets: A deep representation for volumetric shapes.
\newblock In {\em Proceedings of the IEEE conference on computer vision and
  pattern recognition}, pages 1912--1920, 2015.

\bibitem{xu2019deep}
Zexiang Xu, Sai Bi, Kalyan Sunkavalli, Sunil Hadap, Hao Su, and Ravi
  Ramamoorthi.
\newblock Deep view synthesis from sparse photometric images.
\newblock {\em ACM Transactions on Graphics}, 38(4):76, 2019.

\bibitem{xu2018deep}
Zexiang Xu, Kalyan Sunkavalli, Sunil Hadap, and Ravi Ramamoorthi.
\newblock Deep image-based relighting from optimal sparse samples.
\newblock {\em ACM Transactions on Graphics}, 37(4):126, 2018.

\bibitem{yao2018mvsnet}
Yao Yao, Zixin Luo, Shiwei Li, Tian Fang, and Long Quan.
\newblock Mvsnet: Depth inference for unstructured multi-view stereo.
\newblock In {\em Proceedings of the European Conference on Computer Vision
  (ECCV)}, pages 767--783, 2018.

\bibitem{yariv2020multiview}
Lior Yariv, Yoni Kasten, Dror Moran, Meirav Galun, Matan Atzmon, Basri Ronen,
  and Yaron Lipman.
\newblock Multiview neural surface reconstruction by disentangling geometry and
  appearance.
\newblock {\em Advances in Neural Information Processing Systems}, 33, 2020.

\bibitem{zhou2018stereo}
Tinghui Zhou, Richard Tucker, John Flynn, Graham Fyffe, and Noah Snavely.
\newblock Stereo magnification: learning view synthesis using multiplane
  images.
\newblock {\em ACM Transactions on Graphics}, 37(4):1--12, 2018.

\end{thebibliography}


\begin{thebibliography}{1}\itemsep=-1pt

\bibitem{greene1986environment}
Ned Greene.
\newblock Environment mapping and other applications of world projections.
\newblock {\em IEEE Computer Graphics and Applications}, 6(11):21--29, 1986.

\bibitem{mildenhall2020nerf}
Ben Mildenhall, Pratul~P Srinivasan, Matthew Tancik, Jonathan~T Barron, Ravi
  Ramamoorthi, and Ren Ng.
\newblock Nerf: Representing scenes as neural radiance fields for view
  synthesis.
\newblock {\em arXiv preprint arXiv:2003.08934}, 2020.

\bibitem{schoenberger2016mvs}
Johannes~Lutz Sch\"{o}nberger, Enliang Zheng, Marc Pollefeys, and Jan-Michael
  Frahm.
\newblock Pixelwise view selection for unstructured multi-view stereo.
\newblock In {\em European Conference on Computer Vision (ECCV)}, 2016.

\bibitem{sitzmann2019deepvoxels}
Vincent Sitzmann, Justus Thies, Felix Heide, Matthias Nie{\ss}ner, Gordon
  Wetzstein, and Michael Zollhofer.
\newblock Deepvoxels: Learning persistent {3D} feature embeddings.
\newblock In {\em CVPR}, pages 2437--2446, 2019.

\bibitem{sitzmann2019scene}
Vincent Sitzmann, Michael Zollh{\"o}fer, and Gordon Wetzstein.
\newblock Scene representation networks: Continuous 3d-structure-aware neural
  scene representations.
\newblock In {\em Advances in Neural Information Processing Systems}, pages
  1119--1130, 2019.

\end{thebibliography}
}

\end{document}


\newcommand{\netex}{NeuTex}

\title{\netex: Neural Texture Mapping for Volumetric Neural Rendering\\Supplementary Material}

\author{
  {Fanbo Xiang$^1$, Zexiang Xu$^2$, Milo\v{s} Ha\v{s}an$^2$, Yannick Hold-Geoffroy$^2$, Kalyan Sunkavalli$^2$, Hao Su$^1$}\\\\
  {$^1$ University of California, San Diego}\\
  {$^2$ Adobe Research}
}

\maketitle

\newcommand{\Comment}[1]{}

\newcommand{\boldstart}[1]{\noindent\textbf{#1}}
\newcommand{\boldstartspace}[1]{\vspace{0.1in}\noindent\textbf{#1}}

\newcommand{\PixelColor}{\mathbf{I}}
\newcommand{\PixelMask}{M}
\newcommand{\Color}{\mathbf{c}}
\newcommand{\Pos}{\mathbf{x}}
\newcommand{\Dir}{\mathbf{d}}
\newcommand{\UV}{\mathbf{u}}
\newcommand{\Normal}{\mathbf{n}}
\newcommand{\Refl}{\mathbf{r}}

\newcommand{\Trans}{T}
\newcommand{\Dens}{\sigma}
\newcommand{\Step}{\delta}
\newcommand{\Func}{F}
\newcommand{\FuncDens}{F_\Dens}
\newcommand{\FuncUV}{F_\text{uv}}
\newcommand{\FuncInv}{F^{-1}_\text{uv}}
\newcommand{\FuncTex}{F_\text{tex}}
\newcommand{\BlendW}{w}

\section{Cube map Clarification}
\begin{figure}[h]
    \centering
    \includegraphics[width=0.5\linewidth]{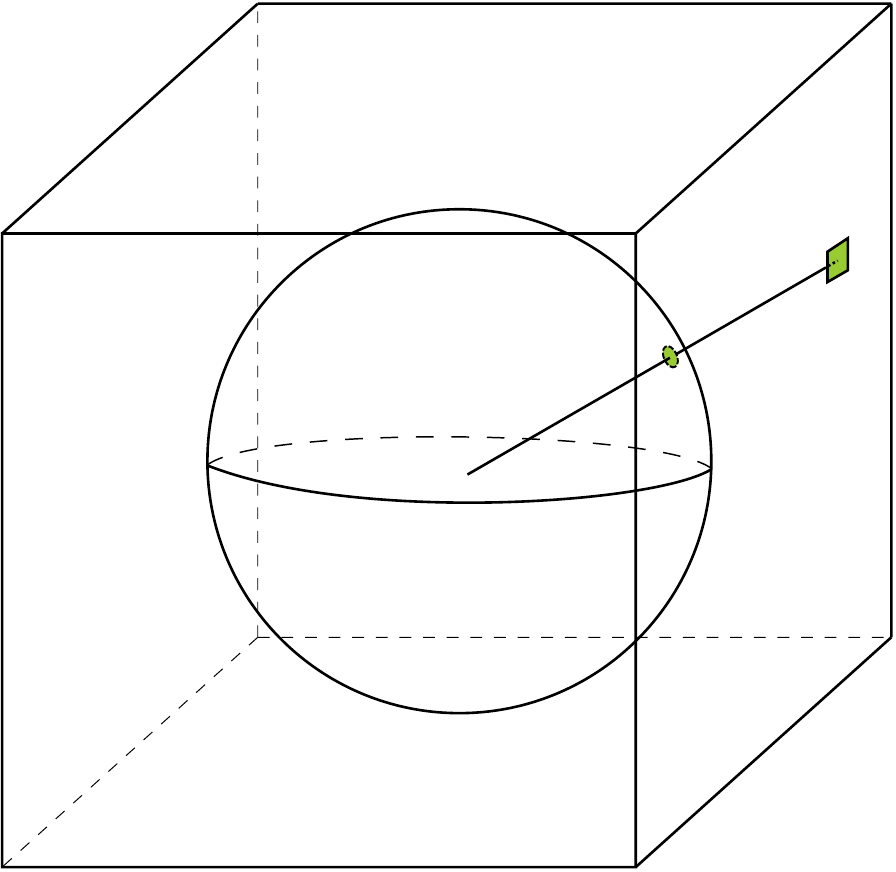}
    \caption{Cube map projection. The color at each point on the unit sphere is projected to a point on the cube centered at the origin. A cubemap is obtained by ``opening up'' the cube.}
    \label{fig:cubemap}
\end{figure}

\begin{figure}[h]
    \centering
    \includegraphics[width=0.8\linewidth]{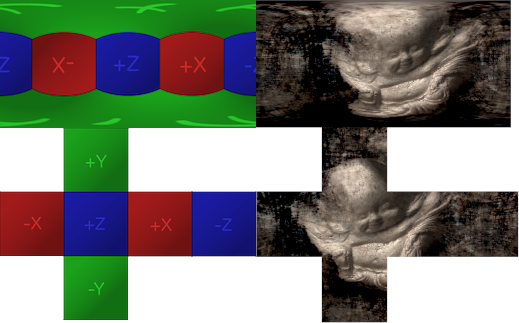}
    \caption{Cube maps on the second row corresponds to the equirectangle maps on the first row. They are different projections of the same spherical texture. A cubemap has a smaller distortion on the Y direction.}
    \label{fig:cubemap2}
\end{figure}

We briefly clarify our texture visualization (with cubemaps) used in our main paper.
As discussed in Sec.~3.3 in the paper, we use spherical UVs for our texture mapping, where $\UV$ represents a point on the surface a unit sphere.
For all the figures in the main paper, we use cubemaps \cite{greene1986environment} to visualize the spherical domain.
A cubemap consists of sixes faces of a unit box, recording all the color information projected from a unit sphere (as shown in Fig~\ref{fig:cubemap}), which is widely used in graphics for spherical mapping.
An alternative standard way to visualize a spherical function is to use a equirectangle map.
We show the correspondence between a cube map and a equirectangle map in Fig.~\ref{fig:cubemap2}.
We use cubemaps in the paper since they involve less distortion, avoiding the distorted regions in the top and bottom of equirectangle maps.

\Comment{
In this work, We use cube maps to visualize spherical textures. For any point on the surface of a unit sphere, we project onto the cube centered at the origin. This projection projects the entire surface of the sphere onto the cube as shown in Figure \ref{fig:cubemap}. Next, we open the surface of the cube onto a plane to obtain the cube map. Figure \ref{fig:cubemap2} shows how cube map is related to the sphere map, another common representation for spherical textures.
}

\section{Network Implementation Details}
\begin{figure*}
    \centering
    \includegraphics[width=0.8\textwidth]{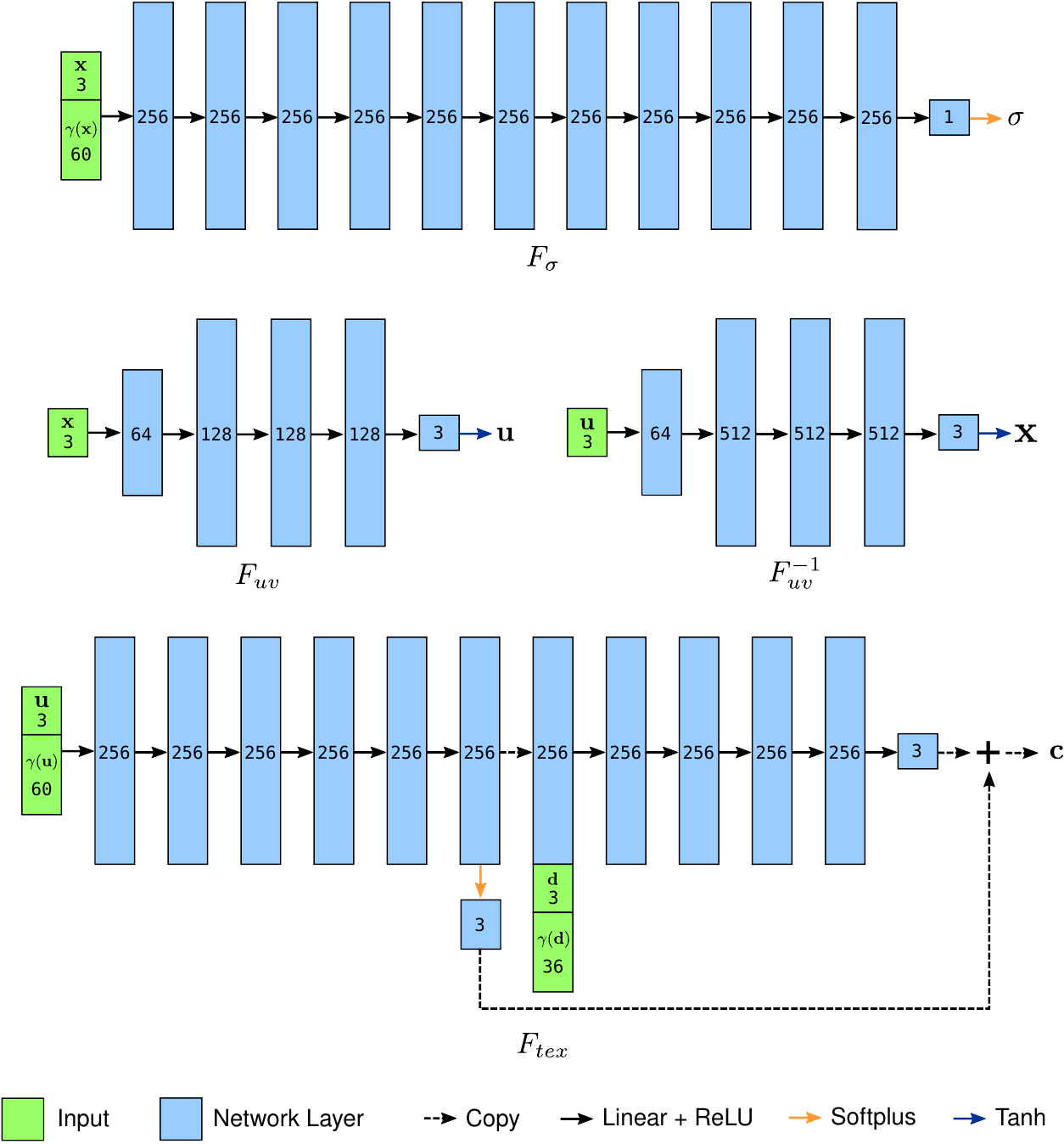}
    \caption{Network structure for the 4 networks. $\textbf{x}$ represents 3D coordinates. $\gamma$ denotes positional encoding. $\textbf{u}$ represents texture-space points (3D points on the unit sphere). $\textbf{d}$ represents a 3D unit vector for viewing direction. $\sigma$ is predicted volumetric density. $\textbf{c}$ is predicted radiance.}
    \label{fig:network}
\end{figure*}
\subsection{Network structure}
We show the detailed network architecture for $F_{\sigma}, F_{uv}, F_{uv}^{-1}$ and $F_{tex}$ in Figure \ref{fig:network}.  

\subsection{Training details in initialization}
Here we describe in detail how we do the initialization stage mentioned in Sec~4.2. in the paper. Given a point cloud from Colmap, we first downsample it to one with 2,000-3,000 points. We denote this point cloud as $P_{gt}$. We then sample 2,500 points uniformly in the UV space (the unit sphere).  We denote the set of UV coordinates as $P$. 

\textbf{Chamfer loss}. The Chamfer loss is simply the Chamfer distance between $F_{uv}^{-1}(P)$ and $P_{gt}$, where $F_{uv}^{-1}(P)$ corresponds to the point cloud generated by inverse-mapping very UV in $P$ to the 3D space using the network $\FuncInv$.
\[L_\text{chamfer} = \text{Chamfer}(F_{uv}^{-1}(P), P_{gt})\]

\textbf{Inverse loss}. We also leverage a loss that is similar to our cycle loss to let the initialization also influence the texture mapping network $\FuncUV$. 
In particular, instead of the 3D-to-2D-to-3D cycle mapping used the cycle loss in Eqn.~12 of the paper, we leverage a 2D-to-3D-to-2D cycle mapping in the initialization, given by:
\[L_\text{cycle2} = || F_{uv}(F_{uv}^{-1}(P)) - P||_2^2\].

\textbf{Rendering and mask loss}. The same rendering and mask loss as described in section 4.1 are also applied in the initialization stage. So the loss at initialization stage is
\[L_\text{init} = L_\text{chamfer} + aL_\text{cycle2} + bL_\text{render} + cL_\text{mask}\]
where we set $a=100$, $b=c=1$.

\section{Additional Results}
\subsection{Full quantitative comparison}
We have shown the averaged quantitative results across five DTU scenes in Tab.~1 of the paper. 
Detailed comparisons on individual scenes are provided in Table \ref{tab:full_number}.
Similar to the average scores, though slightly worse than NeRF, our method significantly outperforms other traditional and neural rendering methods. 

\subsection{Additional visual results}
\begin{figure*}[t]
    \centering
    \begin{minipage}{\textwidth}
    \includegraphics[width=\textwidth]{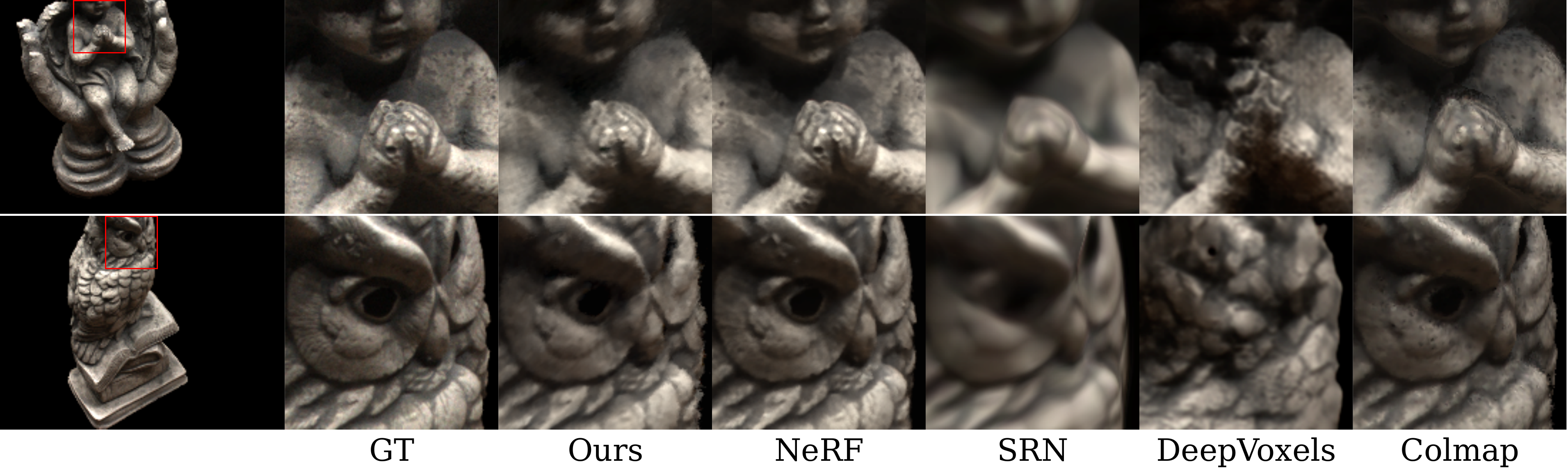}
    \caption{Comparison on the remaining DTU scenes.}
    \label{fig:supp_comparison}
    \vspace{2em}
    \end{minipage}
    \begin{minipage}{\textwidth}
   \centering
    \includegraphics[width=\textwidth]{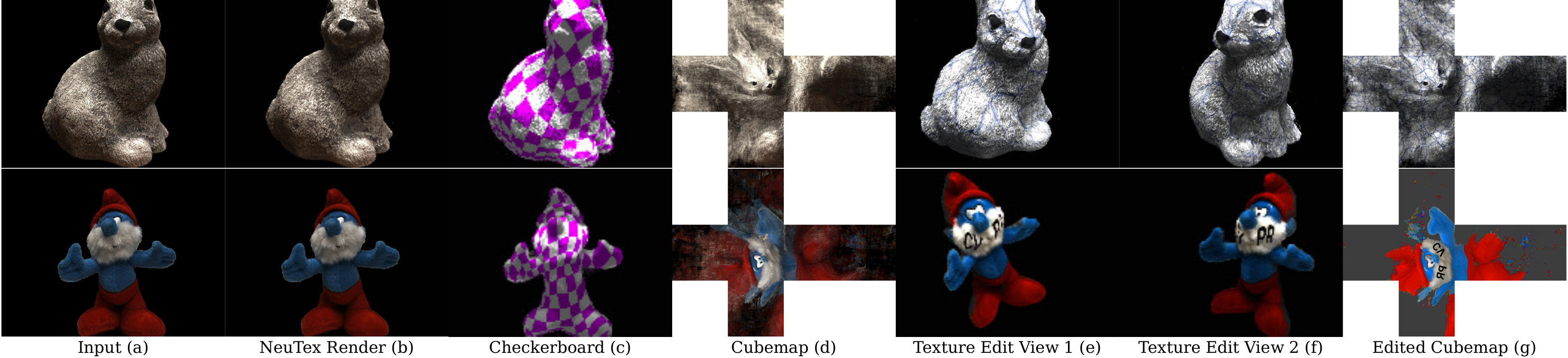}
    \caption{Additional texture editing on DTU scenes.}
    \label{fig:supp_results}
    \end{minipage}%
\end{figure*}

Figure \ref{fig:supp_comparison} shows the visual comparison of different methods on the remaining 2 DTU scenes. Figure \ref{fig:supp_results} shows additional texture editing results. Please refer to the attached video for more results on view synthesis and editing.


\begin{table*}[t]
\centering
\begin{tabular}{l|ll|ll|ll|ll|ll}
 & \multicolumn{2}{c|}{55} & \multicolumn{2}{c|}{83} & \multicolumn{2}{c|}{114} & \multicolumn{2}{c|}{118} & \multicolumn{2}{c}{122} \\ 
Method & PSNR & SSIM & PSNR & SSIM & PSNR & SSIM & PSNR & SSIM & PSNR & SSIM \\\hline
SRN\cite{sitzmann2019scene} & 21.35 & 0.673 & 28.68 & 0.929 & 23.75 & 0.808 & 28.74 & 0.900 & 27.75 & 0.877 \\
DeepVoxels\cite{sitzmann2019deepvoxels} & 17.21 & 0.532 & 23.76 & 0.858 & 17.97 & 0.606 & 23.18 & 0.764 & 22.12 & 0.748 \\
Colmap\cite{schoenberger2016mvs} & 21.25 & 0.784 & 27.11 & 0.921 & 20.69 & 0.809 & 27.43 & 0.907 & 26.66 & 0.905 \\
NeRF\cite{mildenhall2020nerf} & \textbf{26.78} & \textbf{0.913} & \textbf{31.77} & \textbf{0.952} & \textbf{27.38} & \textbf{0.918} & \textbf{33.98} & \textbf{0.954} & \textbf{33.72} & \textbf{0.955} \\
Ours & \textbf{22.67} & \textbf{0.808} & \textbf{30.61} & \textbf{0.931} & \textbf{26.45} & \textbf{0.891} & \textbf{30.67} & \textbf{0.916} & \textbf{30.75} & \textbf{0.925}
\end{tabular}
\caption{PSNR/SSIM for novel view synthesis quality on 4 held-out views on 5 DTU scenes.}
\label{tab:full_number}
\end{table*}

%

{\small
\bibliographystyle{ieee_fullname}
\bibliography{supplement}
}